\documentclass[11pt,a4paper]{article}
\usepackage[hyperref]{acl2020}
\pdfoutput=1
\usepackage{times}
\usepackage{latexsym}

\usepackage{tikz}
\usepackage{todonotes}
\usepackage[ruled,vlined]{algorithm2e}

\usepackage{breakurl}

\usepackage{microtype}

\aclfinalcopy % Uncomment this line for the final submission
\title{Generation of complex database queries and API calls from natural language utterances}

\author{Amol Kelkar, Nachiketa Rajpurohit, Utkarsh Mittal, Peter Relan\\
Got It AI R\&D \\ {\tt \{amol,nachiketa,utkarsh,peter\}@got-it.ai}}
\begin{document}
\maketitle

\begin{abstract}
Generating queries corresponding to natural language questions is a long standing problem. Traditional methods lack language flexibility, while newer sequence-to-sequence models require large amount of data. Schema-agnostic sequence-to-sequence models can be fine-tuned for a specific schema using a small dataset but these models have relatively low accuracy. We present a method that transforms the query generation problem into an intent classification and slot filling problem. This method can work using small datasets. For questions similar to the ones in the training dataset, it produces complex queries with high accuracy. For other questions, it can use a template-based approach or predict query pieces to construct the queries, still at a higher accuracy than sequence-to-sequence models.  On a real-world dataset, a schema fine-tuned state-of-the-art generative model had 60\% exact match accuracy for the query generation task, while our method resulted in 92\% exact match accuracy.
\end{abstract}

\section{Introduction}
Natural language interfaces to databases (NLIDB) \cite{NLI:1987} systems allow a user to communicate with the database directly by entering the query in the form of a natural language question. Pattern matching based \cite{ELIZA:1966}, syntax based \cite{woods1972lunar}, semantic grammar based systems have been investigated \cite{nlidb-review}.

In recent years, neural networks based generative text-2-SQL semantic parsers have gained prominence as a core component in NLIDB systems. They use neural language models that can handle natural language variations and ambiguities better than the traditional approaches. Some of these approaches work with a specific schema, such as Seq2SQL \cite{zhong2017seq2sql}. They require large datasets of question-query pairs for the specific schema, which is typically not available for practical applications. Also, the generated queries are typically simplistic compared to queries used in real world applications. Other approaches such as IRNet \cite{IRNet:Guo2019} and GNN \cite{GNN:Bogin2019} are schema agnostic and accept schema description along with natural language question at inference time. These are trained on a large corpus of question-query pairs across many schemas and are typically fine-tuned on a small dataset for the specific application schema. Unfortunately, their accuracy is not high enough to use in practical applications without human oversight of generated queries.

In this paper, we present an approach that produces complex queries from natural language questions on a specific schema with high accuracy, while requiring only a small dataset of question-query pairs. This approach can be used to produce SQL queries, API calls or other commands.

\section{Our approach}

We use intent discovery, intent classification, entity extraction, slot filling, and template based generation to generate query for a given natural language question.

\subsection{Intent discovery}

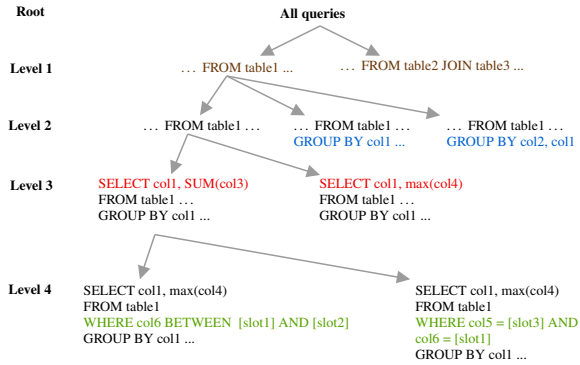
\begin{figure}
\resizebox{1.0\linewidth}{!}{\tikzset{every picture/.style={line width=0.75pt}} %set default line width to 0.75pt        

\begin{tikzpicture}[x=0.75pt,y=0.75pt,yscale=-1,xscale=1]
%uncomment if require: \path (0,285); %set diagram left start at 0, and has height of 285

%Straight Lines [id:da8451552316914623] 
\draw [color={rgb, 255:red, 155; green, 155; blue, 155 }  ,draw opacity=1 ][line width=0.75]    (240,20) -- (202.68,38.66) ;
\draw [shift={(200,40)}, rotate = 333.43] [fill={rgb, 255:red, 155; green, 155; blue, 155 }  ,fill opacity=1 ][line width=0.08]  [draw opacity=0] (8.93,-4.29) -- (0,0) -- (8.93,4.29) -- cycle    ;
%Straight Lines [id:da8695612740235894] 
\draw [color={rgb, 255:red, 155; green, 155; blue, 155 }  ,draw opacity=1 ][line width=0.75]    (240,20) -- (277.32,38.66) ;
\draw [shift={(280,40)}, rotate = 206.57] [fill={rgb, 255:red, 155; green, 155; blue, 155 }  ,fill opacity=1 ][line width=0.08]  [draw opacity=0] (8.93,-4.29) -- (0,0) -- (8.93,4.29) -- cycle    ;
%Straight Lines [id:da38285177694241235] 
\draw [color={rgb, 255:red, 155; green, 155; blue, 155 }  ,draw opacity=1 ][line width=0.75]    (177.86,53.57) -- (161.68,77.51) ;
\draw [shift={(160,80)}, rotate = 304.05] [fill={rgb, 255:red, 155; green, 155; blue, 155 }  ,fill opacity=1 ][line width=0.08]  [draw opacity=0] (8.93,-4.29) -- (0,0) -- (8.93,4.29) -- cycle    ;
%Straight Lines [id:da30515242818621835] 
\draw [color={rgb, 255:red, 155; green, 155; blue, 155 }  ,draw opacity=1 ][line width=0.75]    (177.86,53.57) -- (227.32,78.64) ;
\draw [shift={(230,80)}, rotate = 206.88] [fill={rgb, 255:red, 155; green, 155; blue, 155 }  ,fill opacity=1 ][line width=0.08]  [draw opacity=0] (8.93,-4.29) -- (0,0) -- (8.93,4.29) -- cycle    ;
%Straight Lines [id:da9082806822259689] 
\draw [color={rgb, 255:red, 155; green, 155; blue, 155 }  ,draw opacity=1 ][line width=0.75]    (177.86,53.57) -- (317.05,79.45) ;
\draw [shift={(320,80)}, rotate = 190.53] [fill={rgb, 255:red, 155; green, 155; blue, 155 }  ,fill opacity=1 ][line width=0.08]  [draw opacity=0] (8.93,-4.29) -- (0,0) -- (8.93,4.29) -- cycle    ;
%Straight Lines [id:da9958822711466393] 
\draw [color={rgb, 255:red, 155; green, 155; blue, 155 }  ,draw opacity=1 ][line width=0.75]    (151.86,92.57) -- (141.99,116.79) ;
\draw [shift={(140.86,119.57)}, rotate = 292.17] [fill={rgb, 255:red, 155; green, 155; blue, 155 }  ,fill opacity=1 ][line width=0.08]  [draw opacity=0] (8.93,-4.29) -- (0,0) -- (8.93,4.29) -- cycle    ;
%Straight Lines [id:da5039310437306961] 
\draw [color={rgb, 255:red, 155; green, 155; blue, 155 }  ,draw opacity=1 ][line width=0.75]    (151.86,92.57) -- (237.14,119.11) ;
\draw [shift={(240,120)}, rotate = 197.29] [fill={rgb, 255:red, 155; green, 155; blue, 155 }  ,fill opacity=1 ][line width=0.08]  [draw opacity=0] (8.93,-4.29) -- (0,0) -- (8.93,4.29) -- cycle    ;
%Straight Lines [id:da010853552587912652] 
\draw [color={rgb, 255:red, 155; green, 155; blue, 155 }  ,draw opacity=1 ][line width=0.75]    (130,160) -- (120.95,187.15) ;
\draw [shift={(120,190)}, rotate = 288.43] [fill={rgb, 255:red, 155; green, 155; blue, 155 }  ,fill opacity=1 ][line width=0.08]  [draw opacity=0] (8.93,-4.29) -- (0,0) -- (8.93,4.29) -- cycle    ;
%Straight Lines [id:da0035335705274655016] 
\draw [color={rgb, 255:red, 155; green, 155; blue, 155 }  ,draw opacity=1 ][line width=0.75]    (130,160) -- (297.05,189.48) ;
\draw [shift={(300,190)}, rotate = 190.01] [fill={rgb, 255:red, 155; green, 155; blue, 155 }  ,fill opacity=1 ][line width=0.08]  [draw opacity=0] (8.93,-4.29) -- (0,0) -- (8.93,4.29) -- cycle    ;

% Text Node
\draw (212,6.83) node [anchor=north west][inner sep=0.75pt]  [font=\scriptsize] [align=left] {\textbf{All queries}};
% Text Node
\draw (146,42.83) node [anchor=north west][inner sep=0.75pt]  [font=\scriptsize] [align=left] {\textcolor[rgb]{0.42,0.22,0.04}{… FROM table1 ...}};
% Text Node
\draw (252,41.83) node [anchor=north west][inner sep=0.75pt]  [font=\scriptsize] [align=left] {\textcolor[rgb]{0.42,0.22,0.04}{… FROM table2 JOIN table3 ...}};
% Text Node
\draw (121,81.83) node [anchor=north west][inner sep=0.75pt]  [font=\scriptsize] [align=left] {… FROM table1 …};
% Text Node
\draw (221,81.83) node [anchor=north west][inner sep=0.75pt]  [font=\scriptsize] [align=left] {… FROM table1 … \\\textcolor[rgb]{0,0.38,0.82}{GROUP BY col1 ...}};
% Text Node
\draw (323,81.83) node [anchor=north west][inner sep=0.75pt]  [font=\scriptsize] [align=left] {… FROM table1 … \\\textcolor[rgb]{0,0.38,0.82}{GROUP BY col2, col1}};
% Text Node
\draw (91,120.83) node [anchor=north west][inner sep=0.75pt]  [font=\scriptsize] [align=left] {\textcolor[rgb]{0.93,0,0}{SELECT col1, SUM(col3)} \\FROM table1 … \\GROUP BY col1 ...};
% Text Node
\draw (238,120.83) node [anchor=north west][inner sep=0.75pt]  [font=\scriptsize] [align=left] {\textcolor[rgb]{0.93,0,0}{SELECT col1, max(col4) }\\FROM table1 … \\GROUP BY col1 ...};
% Text Node
\draw (81,192.83) node [anchor=north west][inner sep=0.75pt]  [font=\scriptsize] [align=left] {SELECT col1, max(col4) \\FROM table1 \\\textcolor[rgb]{0.34,0.64,0}{WHERE col6 BETWEEN \ [slot1] AND [slot2] }\\GROUP BY col1 ...};
% Text Node
\draw (302,192.83) node [anchor=north west][inner sep=0.75pt]  [font=\scriptsize] [align=left] {SELECT col1, max(col4) \\FROM table1 \\\textcolor[rgb]{0.34,0.64,0}{WHERE col5 = [slot3] AND}\\\textcolor[rgb]{0.34,0.64,0}{col6 = [slot1]}\\GROUP BY col1 ...};
% Text Node
\draw (36,5.83) node [anchor=north west][inner sep=0.75pt]  [font=\scriptsize] [align=left] {\textbf{Root}};
% Text Node
\draw (32,43.83) node [anchor=north west][inner sep=0.75pt]  [font=\scriptsize] [align=left] {\textbf{Level 1}};
% Text Node
\draw (31,81.83) node [anchor=north west][inner sep=0.75pt]  [font=\scriptsize] [align=left] {\textbf{Level 2}};
% Text Node
\draw (32,121.83) node [anchor=north west][inner sep=0.75pt]  [font=\scriptsize] [align=left] {\textbf{Level 3}};
% Text Node
\draw (31,191.83) node [anchor=north west][inner sep=0.75pt]  [font=\scriptsize] [align=left] {\textbf{Level 4}};

\end{tikzpicture}}
\caption{A set of question-query examples are mapped to a tree of intents by analyzing the queries.}
\label{figure:tree}
\end{figure}

We first analyze queries in examples from the question-query dataset and build a tree of intents. See figure \ref{figure:tree}. Root of the the tree represents all example queries. Each level 1 node represents a subset of queries that uses a specific table and {\small \tt JOIN} set. Each level 2 node represents a further subset that share {\small \tt GROUP BY} columns. Each level 3 node represents a specific set of {\small \tt SELECT} clauses. Each level 4 node represents queries that differ only in the values used.

\subsection{Intent classification and slot filling}

\begin{figure}[t]
\resizebox{1.0\linewidth}{!}{\input{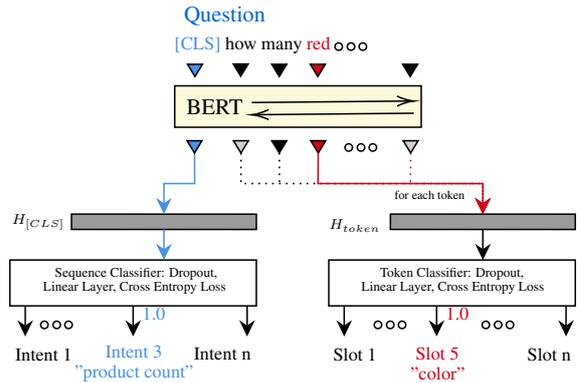}}
\caption{Joint intent detection and slot filling}
\label{figure:idsf}
\end{figure}

We train a BERT \cite{BERT:Devlin2019} fine-tuned joint intent classification and slot filling (IDSF) model. It predicts the intent corresponding to a user question. Each node in the intent tree represents an intent. The model is trained to predict one intent from a specific intent level, or a set of intents along a path from root to a level 4 intent. Figure \ref{figure:idsf} shows configuration where a single intent is predicted. The model is also jointly trained to predict which tokens from the input question represent slot values and which slots should be filled with those values.

\subsection{Query generation}

The slot values predicted by the IDSF model may need to be transformed before use in queries. We build translation dictionaries using annotations from the question-query dataset to map slot values into query values. We also detect slot data types using a heuristic algorithm. Then, for slots with known data types such as date, we detect formatting rules to transform the value into appropriately formatted query value. For example, a date could be specified as {\small \tt "next Thursday"} in the question that gets extracted as a date object, and needs to be used in the query as {\small \tt "2020-10-20"}. In this case, we would detect formatting rule as {\small \tt "\%y-\%m-\%d"}. If the lists of allowed values for certain database columns is available, that information is used to map corresponding slot values to select one of the allowed values.

If the IDSF model predicts a level 4 intent, query can be generated by simply replacing placeholders in the intent's template. If a level 3 intent is predicted, {\small \tt WHERE} and {\small \tt HAVING} conditions in the SQL query are generated based on which slot values are available and rest of the query is generated by simple slot value substitution like level 4 intents. Queries for level 2 and level 1 intents are generated using more complex algorithms and models that predict non-value slots such as aggregation operators and columns to use in {\small \tt SELECT} clause of the query, similar to the approach used in \cite{SQLova:Hwang2019}. If no intent is predicted with high confidence, a generative text-to-SQL model such as \cite{kelkar2020bertranddr} is used, with a human-in-the-loop to validate generated query before use.

The approach can be used to generate commands in an arbitrary non-SQL syntax by pre-processing into an isomorphic SQL command. We use this approach to predict API calls from natural language questions. Consider a question {\small \tt What's the current temperature in Seattle?} and corresponding API call such as {\small \tt temperature(city='Seattle')}. We pre-process the API call as {\small \tt SELECT t FROM temperature WHERE city='Seattle'} during training and back into API call after inference.

\section{Results and conclusion}
We applied this approach to build an NLIDB system for a particular e-commerce application. We used a dataset of 230 question-query pairs and schema description as table-column hierarchy with column types and allowed values for enumerated type columns. 200 examples were used for training and 30 were reserved for testing. We tested GNN \cite{GNN:Bogin2019} trained on the Spider dataset \cite{Spider:Yu2018} and saw 8\% accuracy on the test dataset. When we fine tuned the GNN model on this schema using the training dataset, the accuracy improved to 60\%. In comparison, our system resulted in 92\% accuracy on the test dataset.  

\bibliography{acl2020}
\bibliographystyle{acl_natbib}

\end{document}